\pdfoutput=1

\documentclass[letterpaper]{article} % DO NOT CHANGE THIS
\usepackage{aaai23}  % DO NOT CHANGE THIS
\usepackage{times}  % DO NOT CHANGE THIS
\usepackage{helvet}  % DO NOT CHANGE THIS
\usepackage{courier}  % DO NOT CHANGE THIS
\usepackage[hyphens]{url}  % DO NOT CHANGE THIS
\usepackage{graphicx} % DO NOT CHANGE THIS
\usepackage{natbib}  % DO NOT CHANGE THIS AND DO NOT ADD ANY OPTIONS TO IT
\usepackage{caption} % DO NOT CHANGE THIS AND DO NOT ADD ANY OPTIONS TO IT
\usepackage{algorithm}
\usepackage{algorithmic}

\usepackage{amsmath}
\usepackage{amssymb}
\usepackage{amsfonts}
\usepackage{bm}
\usepackage{multirow}
\usepackage{multicol}
\usepackage{graphicx}
\usepackage{booktabs}
\usepackage{subfigure} 
\usepackage{caption}
\usepackage{balance}
\def\method{GLCC}

\usepackage{newfloat}
\usepackage{listings}
\DeclareCaptionStyle{ruled}{labelfont=normalfont,labelsep=colon,strut=off} % DO NOT CHANGE THIS
\floatstyle{ruled}
\newfloat{listing}{tb}{lst}{}
\floatname{listing}{Listing}
\title{GLCC: A General Framework for Graph-Level Clustering}
\author{
    Wei Ju$^{1,}$\thanks{Equal contribution with order determined by flipping a coin.}, Yiyang Gu$^{1,*}$, Binqi Chen$^{2}$, Gongbo Sun$^{3}$,
    Yifang Qin$^{2}$,\\
    Xingyuming Liu$^{2}$,
    Xiao Luo$^{4,}$\thanks{Corresponding authors.},
    Ming Zhang$^{1,\dagger}$
}
\affiliations{
    $^1$School of Computer Science, Peking University, China\\
    $^2$School of EECS, Peking University, China\\
    $^3$Beijing National Day School, China\\
    $^4$Department of Computer Science, University of California Los Angeles, USA
    \{juwei, yiyanggu, qinyifang, liuxym, mzhang\_cs\}@pku.edu.cn, \\berkinchen@gmail.com, tommy2809@163.com, xiaoluo@cs.ucla.edu
}

\usepackage{bibentry}
\begin{document}

\maketitle

\begin{abstract}
    This paper studies the problem of graph-level clustering, which is a novel yet challenging task. This problem is critical in a variety of real-world applications such as protein clustering and genome analysis in bioinformatics. Recent years have witnessed the success of deep clustering coupled with graph neural networks (GNNs). However, existing methods focus on clustering among nodes given a single graph, while exploring clustering on multiple graphs is still under-explored. In this paper, we propose a general graph-level clustering framework named Graph-Level Contrastive Clustering (\method{}) given multiple graphs. Specifically, \method{} first constructs an adaptive affinity graph to explore instance- and cluster-level contrastive learning (CL). Instance-level CL leverages graph Laplacian based contrastive loss to learn clustering-friendly representations while cluster-level CL captures discriminative cluster representations incorporating neighbor information of each sample. Moreover, we utilize neighbor-aware pseudo-labels to reward the optimization of representation learning. The two steps can be alternatively trained to collaborate and benefit each other. Experiments on a range of well-known datasets demonstrate the superiority of our proposed \method{} over competitive baselines. 
    \end{abstract}

\maketitle

\section{Introduction}

Clustering is a fundamental problem in graph machine learning, which has been widely studied for decades. It aims at partitioning similar samples into the same group and dissimilar samples into different groups. The clusters of samples provide a global insight of the whole dataset, which has various downstream applications, including anomaly detection~\cite{sheng2019multi}, domain adaptation~\cite{tang2020unsupervised}, community detection~\cite{liu2020deep} and representation learning~\cite{xu2021self,luo2022clear}. 

Over the past decades, traditional methods such as spectral clustering~\cite{ng2001spectral} and subspace clustering~\cite{vidal2011subspace} have played a dominant role. However, the separation of representation learning and clustering unavoidably leads to sub-optimal solutions. Due to the strong representation learning capability of deep learning, deep clustering approaches~\cite{caron2018deep,mukherjee2019clustergan,li2021contrastive,zhong2021graph,de2022top} have recently achieved state-of-the-art performance. The crucial characteristic of deep clustering is to learn clustering-friendly representations without manual feature extraction via deep neural networks in an end-to-end fashion. Representative method DeepCluster~\cite{caron2018deep} iteratively groups the features with k-means and uses the cluster assignments as supervision to update the deep neural networks. With deep clustering, representation learning and clustering can be optimized in a joint way to learn clustering-friendly representations.

With the advancement of graph neural networks (GNNs) in achieving unprecedented success for graph-related tasks, one promising direction to leverage GNNs is graph clustering~\cite{bo2020structural,cheng2021multi,peng2021attention,zhao2021graph,pan2021multi,liu2022deep}. The basic idea of graph clustering methods is to train GNNs for learning effective cluster assignments to divide nodes into different groups without human annotations. Specifically, SDCN~\cite{bo2020structural} firstly integrates the structural information into deep clustering combined with autoencoder and GCN. To avoid representation collapse caused by over-smoothing in GCN, DCRN~\cite{liu2022deep} proposes a self-supervised deep graph clustering method by reducing information correlation in a dual manner.

Although existing graph clustering approaches have achieved encouraging performance, they all focus on studying clustering among nodes given a single graph. In other words, they are tailored to node-level clustering. Nevertheless, to the best of our knowledge, clustering on multiple graphs (also called graph-level clustering) remains largely unexplored, and has a variety of real-world applications. For example, protein clustering is a significant topic in bioinformatics, which is used to construct meaningful and stable groups of similar proteins to be used for analysis and functional annotation~\cite{zaslavsky2016clustering}. Moreover, graph-level clustering is a crucial yet challenging task, unlike node-level clustering where we can derive extra supervision for each node from their neighbors via propagation, graphs are individual instances isolated from each other and thus we cannot directly aggregate information from other graphs. 
As such, we are looking for an approach tailored to graph-level clustering that can well learn clustering-friendly representations, and meanwhile capture discriminative clustering information from neighboring graphs.

Towards this end, this work proposes a general framework called Graph-Level Contrastive Clustering (\method{}) on multiple graphs. The key idea of \method{} is to exploit the multi-granularity information to provide a global characterization of graph instances. To achieve this goal effectively, \method{} first constructs an adaptive affinity graph to incorporate neighbor knowledge, we then introduce two-level contrastive learning (CL) based on the affinity graph, i.e., an instance-level CL and a cluster-level CL, respectively. On the one hand, instance-level CL leverages graph Laplacian based contrastive loss to learn clustering-friendly representations for effective cluster assignments. On the other hand, cluster-level CL incorporates neighbor information of each sample to capture compact cluster representations to achieve cluster-level consistency. Furthermore, neighbor-aware pseudo-labels are generated to feed back the training of representation learning, so that the clustering and representation learning can be alternatively optimized to cooperatively supervise and mutually enhance each other. By incorporating this multi-granularity information, our experiments show that it can largely improve the existing state-of-the-art approaches on multiple real-life datasets. To summarize, the main contributions of this work are as follows:

\begin{itemize}
\item \textbf{General Aspects:} 
To the best of our knowledge, this could be the first work to investigate deep graph-level clustering, which explores graph-level clustering on multiple graphs, different from existing works in studying clustering among nodes given a single graph.

\item \textbf{Novel Methodologies:} We propose a general framework to explore instance- and cluster-level contrastive learning based on the affinity graph. Instance-level CL aims to learn clustering-friendly representations, while cluster-level CL captures discriminative cluster representations.
\item \textbf{Multifaceted Experiments:} We conduct comprehensive experiments on various well-known datasets to demonstrate the effectiveness of the proposed approach.

\end{itemize}

\section{Related Work}
\label{sec::related}

\smallskip
\noindent\textbf{Graph Neural Networks.}
GNNs are originally introduced by \cite{gori2005new} and are a typical class of deep neural networks that combine the topological structure and associated features of a graph, thus possessing the powerful capability to process graph-structured data. The basic idea is to learn the low-dimensional graph representations through a recursive neighborhood aggregation scheme~\cite{gilmer2017neural,ju2022kgnn,luo2022dualgraph}. The derived graph representations can be used to serve various downstream tasks, such as node classification~\cite{kipf2017semi}, graph classification~\cite{ju2022tgnn}, and graph clustering~\cite{bo2020structural}.

\smallskip
\noindent\textbf{Deep Clustering.}
Our work is related to deep clustering, which has achieved impressive performance, benefiting from the breakthroughs in deep learning. There has been a surge of interest in employing deep neural networks to enhance clustering, which can be divided into two main categories: (i) reconstruction based methods, and (ii) self-augmentation based methods. For the first category, it aims to leverage the auto-encoder~\cite{rumelhart1985learning} paradigm to impose desired constraints on feature learning in the latent space. For example, Deep Embedded Clustering
(DEC)~\cite{xie2016unsupervised} simultaneously learns feature representations and cluster assignments by minimizing the Kullback-Leibler divergence. IDEC~\cite{guo2017improved} improves the clustering by preserving the local structure of data generating distribution. For the second category, the underlying concept is training the networks to achieve the consistency between original samples and their augmented samples. For instance, IIC~\cite{ji2019invariant} maximizes the mutual information of paired dataset samples to keep a consistent assignment probability. However, these methods are not tailored to graph-level clustering, and show an inability to process complex data structures, such as graph domains.

\smallskip
\noindent\textbf{Graph Clustering.}
Another category of related work is graph clustering. Benefiting from the strong capability of GNNs in incorporating both node attributes and graph structures, GNNs have emerged as a powerful approach for graph clustering, which aims to reveal the underlying graph structure and divides the nodes into several disjoint groups. Similarly, most existing graph clustering approaches~\cite{wang2017mgae,wang2019attributed,pan2019learning,fan2020one2multi,bo2020structural} also follow the framework of auto-encoder, in which the graph auto-encoder (GAE) and the variational graph auto-encoder (VGAE) are used to learn the graph-structured data. For example, DAEGC~\cite{wang2019attributed} utilizes an attention network to capture the importance of the neighboring nodes, and further encodes the topological structures and node contents to a compact representation based on GAE. The adversarially regularized graph autoencoder (ARGA)~\cite{pan2019learning} enhances the clustering via introducing an adversarial learning scheme to learn the graph embedding. Compared with existing methods for node-level clustering, our work goes further and studies an under-explored yet important graph-level clustering.

\begin{figure*}[t!]
    \centering
    \includegraphics[width=0.94\textwidth]{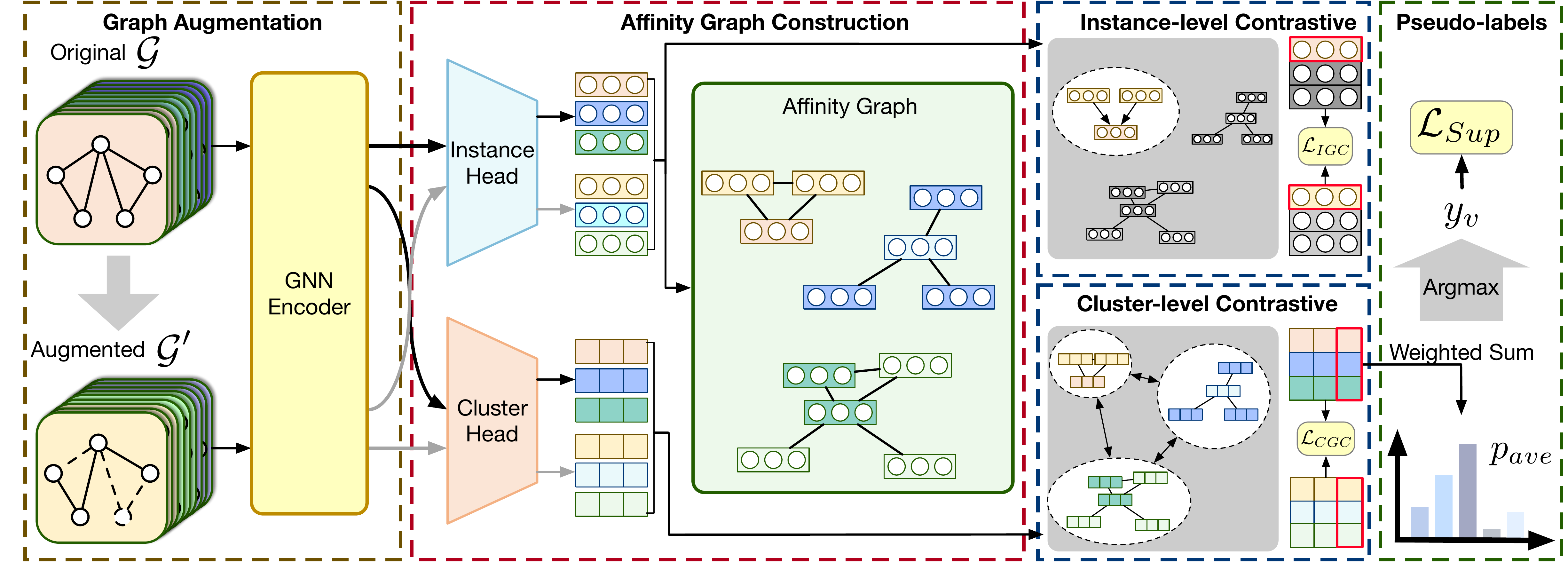}
    \caption{{Illustration of the proposed framework \method{}. 
    }}
    \label{fig:framework}
\end{figure*}

\section{Problem Definition \& Preliminary}

\smallskip
\noindent\textbf{Definition: Graph-level Clustering. } A graph is denoted as a topological graph $G=(\mathcal{V}, \mathcal{E}, \mathbf{X})$, where $\mathcal{V}$ is the set of nodes, $\mathcal{E}$ is the set of edges. We use $\mathbf{X}\in R^{|\mathcal{V}|\times d}$ to denote the node feature matrix, where $d$ is the dimension of features. Let $\mathcal{G} = \left\{G_{1}, \cdots, G_{N}\right\}$ denote a set of unlabeled graphs from $K$ different categories. The goal of graph-level clustering is to separate these graphs into $K$ different clusters such that the graphs with the same semantic labels can be grouped into the same cluster.

\subsection{Graph Neural Networks}
\label{sec::gnn}

We build upon graph neural networks (GNNs) to learn effective graph-level representations. GNNs are powerful architectures by iteratively aggregating over local node neighborhoods via message-passing mechanisms~\cite{gilmer2017neural}. Formally, at $l$-th layer of GNNs, the representation of node $v$ is updated as follows:
\begin{equation}
\label{eq:gnn}
    h_{v}^{(l)}=\mathcal{U}^{(l)}\left(h_{v}^{(l-1)}, \mathcal{A}^{(l)}\left(\left\{h_{u}^{(l-1)}\right\}_{u \in \mathcal{N}(v)}\right)\right),
\end{equation}
where $\mathcal{N}(v)$ are neighbors to node $v$, $\mathcal{U}$ and $\mathcal{A}$ denote the updating and aggregation functions. After $L$ layers of iteration, we combine all node representations to generate the whole graph representation $h_{G}$:
\begin{equation}
\label{eq:readout}
h_{G} = \text{READOUT} (\{ h_{v}^{L} : v \in
\mathcal{V} \}),
\end{equation}
where $\text{READOUT}$ function can be achieved by directly using average, sum, or some adaptive pooling function.

\section{Methodology}
\label{sec::model}

\subsection{Overview}
\label{sec::overview}

This paper proposes a general framework \method{} for graph-level clustering as shown in Figure \ref{fig:framework}. At a high level, \method{} aims to leverage the multi-granularity information to provide a global characterization of graph instances for effective clustering. Specifically, \method{} first constructs an adaptive affinity graph to link semantically similar samples, and then introduces two-level contrastive learning based on the affinity graph. On the one hand, \method{} conducts contrastive learning coupled with graph Laplacian to learn clustering-friendly representations from the instance-level view. On the other hand, \method{} encourages the consistency between a sample and its neighbors to capture compact cluster representations from the cluster-level view. Moreover, predicted neighbor-aware pseudo-labels are used to optimize the representation learning process instead. The two steps can be alternatively trained to collaborate with each other.

\subsection{Adaptive Affinity Graph Construction}
\label{sec::affinity_graph}

A fundamental perception is that similar graphs will tend to involve similar properties. For example, in the biochemistry domain, hexane and heptane that share the same functional groups are prone to exhibit similar boiling points and water solubility, thus easily assigned to the same cluster group. Also, the relationship among graphs changes across properties, showing different clustering attributions. Naturally, how to capture the relationship between different graphs is a key factor affecting the clustering performance.

As such, we further propose an adaptive affinity graph learning module to capture the relationship among different graphs, such that the cluster information can be efficiently propagated between similar graphs to incorporate prior neighbor knowledge.

Formally, we construct an affinity graph $G_A=(\mathcal{V}_G, \mathcal{E}_G)$, which is an undirected graph consisting of $N$ graphs. Denote $\mathcal{V}_G=\{v_{G_1},\cdots, v_{G_N}\}$ the set of nodes, and $\mathcal{E}$ the set of edges represented by the adjacency matrix $A^{(t)}$ such that:
\begin{equation}
\label{eq:knn}
    A_{ij}= \begin{cases}e^{h_{G_i}\cdot h_{G_j}/\tau}, & \text { if } h_{G_j}\in \mathcal{N}^{k}(h_{G_i}) \\ 0, & \text { otherwise }\end{cases}
\end{equation}
where $i,j=1,\cdots,N$, and $\tau$ is a temperature parameter. The $\cdot$ symbol denotes the inner product. $\mathcal{N}^{k}(z)$ are $k$ neighbors to the node representation $z$.
Therein, the edge weight is defined as the similarity between two neighbor nodes.

In this way, the adaptive affinity graph can be progressively updated as the learning process, such that the connecting graphs have similar properties and thus provide richer prior knowledge for clustering. Moreover, the inclusion of the affinity graph overcomes the deficiency of the independence of multiple graphs, and each sample can aggregate and derive extra supervision signals from neighboring graphs, better serving the downstream clustering task.

\subsection{Instance-level Graph Contrast}

There is a key observation that the rows of the feature matrix (i.e., instance representations of graphs) could be interpreted as the clustering assignment probabilities when representation is projected into a $K$-dimensional embedding. Therefore, an effective representation is beneficial to the clustering process. In view of this, motivated by the prominent success of contrastive learning, which possesses the powerful capability of learning discriminative representations from data themselves, we resort to this technique for better representation learning. The basic idea of contrastive learning is to transform the data to generate two augmented views, and compare pairs of instance representations to push away representations from different samples while pulling together those from transformations of the same sample.

To better draw together intra-cluster sample pairs and avoid false negative samples in contrastive learning, we propose a novel graph Laplacian based contrastive learning coupled with our adaptive affinity graph. 

Technically, let $D$ denote the diagonal degree matrix of the adjacency matrix $A$ of the affinity graph, in which $d_{i}$ represents the degree of node $v_{G_i}$, then the normalized symmetric graph Laplacian of $G_A$ can be defined as:
\begin{equation}
\label{eq:laplacian}
L = I - D^{-\frac{1}{2}}AD^{-\frac{1}{2}},
\end{equation}
where $L_{ij} = -\frac{A_{ij}}{\sqrt{d_{i}d_{j}}}, i\neq j$.

Given $N$ original features $\mathbf{H} = \{h_{1}, ..., h_{N}\}$ derived from GNNs followed by a instance head (i.e., multi-layer perceptron, MLP), and the augmented features $\mathbf{H'} = \{h'_{1}, ..., h'_{N}\}$ by randomly selecting one of the four graph augmentations in \cite{you2020graph}, we expect that the original node and the augmented neighbor node that are adjacent on the affinity graph should be pulled closer, while non-adjacent nodes should be pushed away. In other words, $h_{i}$ should be close to $h'_{j}$ if $A_{ij} > 0$ while $h_{i}$ should be far away from $h'_{j}$ if $A_{ij} = 0$. Assume that the dataset can be grouped into several clusters, and we aim to increase the similarities of intra-cluster and decrease those of inter-cluster.

Formally, we can define
\begin{equation}
% \begin{aligned}
    \mathcal{S}_{intra} = \sum_{L_{ij} < 0}-L_{ij}S(h_{i}, h'_{j}),
    \mathcal{S}_{inter} = \sum_{L_{ij} = 0}S(h_{i}, h'_{j})
% \end{aligned}
\end{equation}
as the total similarities of intra-cluster and inter-cluster respectively, where $S(h_{i}, h'_{j})$ is defined in Eq.~\eqref{eq:knn}. Then, our instance-level graph contrastive loss can be defined as:
\begin{equation}
\label{eq:igc_loss}
    \mathcal{L}_{IGC} = -\frac{1}{N}\sum_{i=1}^{N}\log\left(\frac{\sum_{L_{ij}<0}-L_{ij}e^{h_{i}\cdot h_{j}^{'}/\tau}}{\sum_{L_{ij}=0}e^{h_{i}\cdot h_{j}^{'}/\tau}}\right).
\end{equation}

As such, minimizing $\mathcal{L}_{IGC}$ can achieve the goal that samples in the same cluster are more similar in some sense to each other than to those in other clusters and further improve the separableness of the whole dataset.

\subsection{Cluster-level Graph Contrast}

Apart from the instance representations derived from the rows of the feature matrix, correspondingly, the columns of the feature matrix could be interpreted as the clustering representations. In other words, when we project the feature matrix into a subspace whose dimensionality of the rows equals the number of clusters, the columns can be treated as the cluster distributions over instances. 

In this way, conducting contrastive learning on clustering representations can make the different cluster information well capture the inherent characteristics. Nevertheless, traditional contrastive learning at the cluster-level solely treats graphs as individual instances, and hence fails to establish the relationship between the dimensions of the clustering representations. Inspired by this motivation, we leverage the affinity graph to satisfy that each node and its neighbors should be closer with a similar cluster assignment.

Technically, we encourage that the original node should be similar to the augmented view of the neighbor nodes that are adjacent on the affinity graph. Similarly, consider the projected feature matrix with a dimensionality of the cluster number $K$, which is derived from GNNs followed by a cluster head (i.e., MLP), then the columns of the projected feature matrix are denoted as $
\mathbf{Z}={[z_{1}, ..., z_{K}]}_{N\times K}$, and accordingly the column vectors of the augmented neighbor matrix (by randomly sampling a neighbor of each node) are $\tilde{\mathbf{Z}}^{'}={[\tilde{z}_{1}^{'}, ..., \tilde{z}_{K}^{'}]}_{N\times K}$,
where the $i$-th column of the feature matrix can be seen as a representation of the $i$-th cluster, and the intuition is that all columns should differ from each other. Hence we can adopt the idea of contrastive learning to define our cluster-level graph contrastive loss as:
\begin{equation}
\label{eq:cgc_loss}
    \mathcal{L}_{CGC} = -\frac{1}{K}\sum_{i=1}^{K}\log\left(\frac{e^{z_{i}\cdot \tilde{z}_{i}^{'}/\tau}}{\sum_{j=1}^{K}e^{z_{i}\cdot \tilde{z}_{j}^{'}/\tau}}\right)- H(\mathbf{Z}),
\end{equation}
where $H(\cdot)$ is the entropy function to prevent collapsing into trivial outputs of the same cluster.

\subsection{Neighbor-aware Pseudo Labeling}

Note that although discriminative representations can lead to better cluster assignments, representation learning and clustering processes should collaborate and mutually enhance each other. Towards this end, we propose to leverage the confidence-based pseudo-labels generated by cluster assignments to optimize representation learning instead.

Actually, we find that the well-trained model is prone to make predictions with high confidence. In other words, the predicted probability distribution approaches a one-hot vector. However, experiments show that these predictions have low accuracy, making it hard to directly select the confident pseudo-labels. To that effect, we overcome these challenges by incorporating neighbor information from the affinity graph. We believe that the cluster assignments of neighbors could serve and rectify the biased pseudo-labels. 

Specifically, we combine the neighbors' cluster assignments and the target sample itself $p_{v}$, and adopt the operation of weighted summation to calculate the averaged assignment distribution $p_{ave}$, then the hard label of the target sample $y_{v}$ is assigned by the maximum value of the averaged assignment probability:
\begin{equation}
\label{eq:pseudo_labels}
    \begin{aligned}
    &p_{ave}= p_v + \sum\nolimits_{u \in \mathcal{N}(v)} a_{vu} \cdot p_u\\
    &y_{v}=\arg \max \left(p_{ave}\right),
    \end{aligned}
\end{equation}
where $p_{v}$ denotes the $v$-th row of the projected feature matrix with the dimensionality of $K$, the weight $a_{vu}$  is the normalization of the edge weight $A_{vu}$ of the sample $v$ and its neighbor $v$. Finally, top-$\lfloor rN \rfloor$ samples with the minimum entropy value of assignment distribution $p_{ave}$ are selected as subsets to optimize representation learning, where $r$ is the pseudo-label ratio for each dataset.

Once we have the pseudo-labels, we can in turn provide supervision signals for representation learning. Let $i\in I=\{1,...,2\lfloor rN \rfloor\}$ be the index of $\lfloor rN \rfloor$ samples and their augmentations. Here we adopt supervised contrastive learning~\cite{khosla2020supervised}:
\begin{equation}
\label{eq:supervised_loss}
    \mathcal{L}_{Sup}
    =\sum_{i\in I}\frac{-1}{|Q(i)|}\sum_{j\in Q(i)}\log{\frac{\exp\left(h_i\cdot h_j/\tau\right)}{\sum\limits_{a\in A(i)}\exp\left(h_i\cdot h_a/\tau\right)}},
\end{equation}
where $A(i)= I\setminus\{i\}$, $Q(i)=\{q\in A(i):y_q=y_i\}$ is the set of indices of all positives distinct from $i$.

\subsection{Training and Optimization} 

\begin{algorithm}[tb]
    \caption{Optimization Algorithm of \method{}}
    \label{alg:algorithm}
    \textbf{Input}: Training graphs $\mathcal{G} = \left\{G_{1}, \cdots, G_{N}\right\}$, number of clusters $K$, and temperature parameter $\tau$\\
    % \textbf{Parameter}: Optional list of parameters\\
    \textbf{Output}: Cluster assignments

    \begin{algorithmic}[1] %[1] enables line numbers
    \STATE Initializing affinity graph $A$ and GNN parameter $\theta$.\\
    \WHILE{not done}
    \STATE Sample a mini-batch from $\mathcal{G}$ and their neighbors according to $A$.
    \STATE Sample one augmentation from \cite{you2020graph}.
    \STATE \textrm{// \emph{Step 1}}\\
    \STATE Compute instance-level graph contrastive loss $\mathcal{L}_{IGC}$ by Eq.~\eqref{eq:igc_loss}.
    \STATE Compute cluster-level graph contrastive loss $\mathcal{L}_{CGC}$ by Eq.~\eqref{eq:cgc_loss}.
    \STATE Update parameter $\theta$ by gradient descent to minimize $\mathcal{L}_{GC}$ by Eq.~\eqref{eq:gc_loss}.
    \STATE \textrm{// \emph{Step 2}}\\
    \STATE Generate pseudo-labels by Eq.~\eqref{eq:pseudo_labels}.
    \STATE Update parameter $\theta$ by gradient descent to optimize instance representations by Eq.~\eqref{eq:supervised_loss}.
    \STATE \textrm{// \emph{Refine affinity graph}}\\
    \STATE Update $A$ according to Eq.~\eqref{eq:knn}.
    \ENDWHILE
    \end{algorithmic}
\end{algorithm}

To effectively train our framework, we alternatively optimize the following two steps until convergence.

\smallskip
\textbf{Step 1.}
We jointly optimize instance- and cluster-level contrastive learning, the overall objective can be written as:
\begin{equation}
\label{eq:gc_loss}
    \mathcal{L}_{GC} = \mathcal{L}_{IGC} + \mathcal{L}_{CGC}.
\end{equation}

\smallskip
\textbf{Step 2.}
Then we depend on the generated neighbor-aware pseudo-labels to optimize the instance representations through Eq.~\eqref{eq:supervised_loss}.

After each iteration, we can utilize the learned instance representations to refine the adaptive affinity graph, making it progressive to involve more abundant neighbor knowledge, which can better serve the clustering.

\section{Experiments}

\subsection{Experimental Setup}

\noindent\textbf{Datasets.}
We conduct extensive experiments on two kinds of datasets: biochemical molecule datasets and social network datasets. For biochemical molecule datasets, we adopt DD from TU datasets \cite{morris2020tudataset}, and AnchorQuery collected from AnchorQuery platform\footnote{http://anchorquery.csb.pitt.edu/} to test clustering performance on large cluster number. 
% AnchorQuery platform combines the concept of anchors with multicomponent reaction (MCR) \cite{weber2002multi,domling2012chemistry} to target protein–protein interactions, which is important yet difficult in drug discovery. 
% Every anchor-oriented library in AnchorQuery contains compounds generated from different MCR reactions. 
Specifically, we construct AnchorQuery-10K and AnchorQuery-25K datasets with compounds generated from $10$ and $25$ types of multicomponent reactions (MCRs), respectively. The goal is to distinguish reaction types of compounds through graph-level clustering. For social network datasets, we adopt IMDB-B, REDDIT-B, and REDDIT-12K datasets from TU datasets. Dataset statistics are reported in Table~\ref{tab:statistics}.

\begin{table}[!t]
\centering
\tabcolsep=3pt
\resizebox{0.475\textwidth}{!}{
\begin{tabular}{c  c  c  c  c  c } 

\toprule
  Datasets & Category & $\#$Class & $\#$Graph & $\#$Node & $\#$Edge \\

\midrule
  DD & Molecules  & 2 & 1178 & 284.32 & 715.66 \\
  AnchorQuery-10K & Molecules  & 10 & 14774 & 39.80 & 86.69 \\
  AnchorQuery-25K & Molecules  & 25 & 30692 & 39.45 & 86.30 \\

\midrule
IMDB-B & Social Networks  & 2 & 1000 & 19.77 & 96.53 \\
  REDDIT-B & Social Networks  & 2 & 2000 & 429.63 & 497.75 \\

  REDDIT-12K & Social Networks  & 11 & 11929 & 391.41 & 456.89 \\

\bottomrule
\end{tabular}}
\caption{Statistics of datasets.}
\label{tab:statistics}
\end{table}

\begin{table*}[t]
	\centering
        \tabcolsep=4.5pt
	\resizebox{1.01\textwidth}{!}{
	\begin{tabular}{@{}lcccccccccccccccccc@{}}
	\toprule
	Dataset &
	  \multicolumn{3}{c}{DD} &
	  \multicolumn{3}{c}{AnchorQuery-10K} &
	  \multicolumn{3}{c}{AnchorQuery-25K} &
	  \multicolumn{3}{c}{IMDB-B} &
	  \multicolumn{3}{c}{REDDIT-B} &
	  \multicolumn{3}{c}{REDDIT-12K} \\ \midrule
	Metrics &
	  \multicolumn{1}{l}{NMI} &
	  \multicolumn{1}{l}{ACC} &
	  \multicolumn{1}{l}{ARI} &
	  \multicolumn{1}{l}{NMI} &
	  \multicolumn{1}{l}{ACC} &
	  \multicolumn{1}{l}{ARI} &
	  \multicolumn{1}{l}{NMI} &
	  \multicolumn{1}{l}{ACC} &
	  \multicolumn{1}{l}{ARI} &
	  \multicolumn{1}{l}{NMI} &
	  \multicolumn{1}{l}{ACC} &
	  \multicolumn{1}{l}{ARI} &
	  \multicolumn{1}{l}{NMI} &
	  \multicolumn{1}{l}{ACC} &
	  \multicolumn{1}{l}{ARI} &
	  \multicolumn{1}{l}{NMI} &
	  \multicolumn{1}{l}{ACC} &
	  \multicolumn{1}{l}{ARI} \\ \midrule
	Graphlet    & 0.004 & 0.579 & -0.001 &0.053& 0.192&	0.025 & 0.01 & 0.136 & 0.009 & 0.020 & \underline{0.583} & 0.026 & 0.001 & 0.502 & 0.000 & 0.073 & 0.187 & -0.004 \\
	SP         & 0.003 & 0.585 & 0.001 & 0.048& 0.169&	0.016& 0.152 & 0.156 & 0.050 & 0.035 & 0.567 & \underline{0.044} & 0.021 & \underline{0.577} & \underline{0.022} & 0.062 & 0.204 & 0.005 \\
	WL         & 0.006 & 0.586 & 0.002  &0.033& 0.162	&0.010& 0.159 & 0.163 & 0.068 & 0.023 & 0.53 & 0.003 & \underline{0.089} & 0.576 & 0.021 & 0.092 & 0.189 & \underline{0.044}  \\ \midrule
	InfoGraph        & 0.008 & 0.558 & -0.006 & 0.072&0.238&	0.036 & 0.178 & 0.181 & 0.061 & 0.041 & 0.538 & 0.005 & 0.016 & 0.508 & 0.000 & 0.045 & 0.205 & 0.003 \\
	GraphCL         & \underline{0.019} & 0.573 & -0.009 & \underline{0.074}& \underline{0.239}	&0.037& 0.195 & 0.201 & 0.074 & 0.046 & 0.545 & 0.008 & 0.033 & 0.519 & 0.001 & \underline{0.096} & 0.181 & 0.021 \\
	CuCo        & 0.012 & 0.562 & -0.010 & 0.072&0.238&	\underline{0.038}&	 0.194 & 0.189 & 0.073 & 0.001 & 0.507 & 0.000 & 0.018 & 0.510  & 0.000 & 0.003	& 0.192 &	0.002   \\
	JOAO      & 0.012&	0.578&	-0.004	&0.069& 0.235& 0.033&		\underline{0.197} &	\underline{0.205}&	\underline{0.076}&	0.042&	0.543&	0.008&	0.034&	0.520&	0.001&	0.003&	0.183&	0.001  \\
	RGCL      & 0.014	&0.565&	-0.009&0.063&0.214	&0.028  &			0.190&	0.182&	0.059&	0.047&	0.546&	0.007&	0.017&	0.509&	0.001&	0.003&	0.092&	0.001  \\
	SimGRACE        & 0.001 & \underline{0.589}	& \underline{0.003}	&0.068&	0.226	&0.031	&	0.189 &	0.186	&0.074&	\underline{0.049}&	0.559&	0.007&	0.024&	0.513&	0.001&	0.062&	\underline{0.210}&	0.005  \\
	
	\midrule
	\textbf{Ours} & \textbf{0.024} & \textbf{0.607} & \textbf{0.023} & \textbf{0.076} & \textbf{0.247} & \textbf{0.043} & \textbf{0.209} & \textbf{0.228} & \textbf{0.083} & \textbf{0.081} & \textbf{0.665} & \textbf{0.106} & \textbf{0.092} & \textbf{0.676} & \textbf{0.087} & \textbf{0.105} & \textbf{0.226} & \textbf{0.058} \\ \bottomrule
	\end{tabular}
	}
	\caption{The clustering performance on six graph property prediction benchmarks. The best results are shown in boldface.}
	\label{tab:result}
	\end{table*}

\smallskip
\noindent\textbf{Evaluation Metrics.}
We adopt three commonly used metrics to evaluate the clustering performance, including Normalized Mutual Information (NMI)~\cite{strehl2002cluster}, clustering Accuracy (ACC)~\cite{li2006relationships} and Adjusted Rand Index (ARI)~\cite{hubert1985comparing}. They test different aspects of the clustering results. NMI and ACC range in $[0,1]$, while ARI ranges in $[-1,1]$. The larger value reflects the better performance for all three metrics.

\smallskip
\noindent\textbf{Baseline Methods.}
We compare our \method{} with two families of baselines:
graph kernel methods and graph contrastive learning methods. The graph kernel methods include Graphlet Kernel \cite{shervashidze2009efficient}, Shortest Path (SP) Kernel \cite{borgwardt2005shortest} and Weisfeiler-Lehman (WL) Kernel \cite{shervashidze2011weisfeiler}. While graph contrastive learning methods include InfoGraph \cite{sun2020infograph}, GraphCL \cite{you2020graph}, CuCo \cite{chu2021cuco}, JOAO \cite{you2021graph}, RGCL \cite{li2022let} and SimGRACE \cite{xia2022simgrace}. These baselines first learn graph representations and then utilize K-means \cite{macqueen1967classification} to cluster instances based on graph representations. Note that there is a lack of graph-level clustering methods for joint learning of representation learning and clustering.

\smallskip
\noindent\textbf{Implementation Details.}
For a fair comparison with previous graph contrastive learning methods, we adopt Graph Isomorphism Network (GIN) \cite{xu2019powerful} as the backbone for all baselines. The number of GIN layers is $3$, and the hidden dimension is set to $64$. The batch size is set to $64$ for DD, IMDB-B, and REDDIT-B, and $256$ for AnchorQuery-10K, AnchorQuery-25K, and REDDIT-12K. 
The temperatures in instance- and cluster-level graph contrastive losses are set to $0.1$ and $1.0$, respectively. For affinity graph construction, we set neighbor number $k = 5$. Pseudo-label ratio $r$ is set to $0.1$. The perturbation ratio of graph augmentation is set to $0.1$. The total number of training epochs is set to $100$.

\subsection{Experimental Results}
We report the quantitative results of our approach against competitive clustering methods in Table \ref{tab:result}. According to the results, we make the following observations: 

\begin{itemize}
\item Overall, our framework \method{} achieves the best performance against other graph kernel methods and graph contrastive learning methods on all six datasets. In particular, \method{} outperforms the closest competitor on IMDB-B with $8.2\%$ and REDDIT-B with $9.9\%$, in terms of ACC, which demonstrates the remarkable capability of our framework for graph-level clustering.

\item Kernel methods generally perform worse than graph contrastive learning methods on biomedical datasets, but the situation is reversed on social datasets. Maybe the reason is that traditional graph kernels are difficult to capture the functional groups of molecules via handcraft substructures, but excel at exploring the path information of relationship connections in social network datasets.

\item The performance of graph contrastive learning methods is generally inferior to our \method{} on IMDB-B and REDDIT-B datasets, which suggests that only instance-level contrastive learning fails to learn effective representations for clustering. In other words, cluster-level contrastive learning is essential to graph-level clustering.

\item For datasets of large cluster numbers, like REDDIT-12K and AnchorQuery-25K, \method{} also shows the superiority over all the strong baselines, which proves the robustness of our framework to various cluster numbers.

\end{itemize}

\subsection{Ablation Study}

We investigate the effectiveness of model components from three aspects: instance- and cluster-level contrast, adaptive affinity graph, and neighbor-aware pseudo label. The results are summarized in Table \ref{tab:ab}.

\smallskip
\noindent\textbf{Effect of Instance- and Cluster-level Contrast.}
$M_1$ only uses instance-level graph contrast, which is equivalent to GraphCL~\cite{you2020graph}. $M_2$ only utilizes cluster-level graph contrast to learn cluster assignments. While $M_3$ combines instance- and cluster-level contrastive learning. We can see that either absence of instance-level contrast or cluster-level contrast will damage the cluster performance, indicating that graph representation learning and clustering can promote and benefit each other.

% Ablation Table
\begin{table}[!t] 
\footnotesize
\tabcolsep=4.5pt
\resizebox{0.48\textwidth}{!}{
\centering
	\begin{tabular}{@{\extracolsep{-0.1cm}}c|cccc|ccc|ccc@{}}
% 		\hline
% 		\hline
\toprule
		\multirow{2}{*}{} & \multicolumn{4}{c|}{Correlations} & \multicolumn{3}{c|}{IMDB-B} & \multicolumn{3}{c}{AnchorQuery-25K} \\
		& IGC & CGC & AAG & NPL  & NMI & ACC & ARI & NMI & ACC & ARI \\
% 		\hline
% 		\hline
\midrule
		$M_1$ & $\surd$ &  &  &  &  0.046	& 0.545 &	0.008 &	0.195 &	0.201 &	0.074   \\
		$M_2$ & & $\surd$ &  &   & 0.059 &	0.631 &	0.071 &	0.191 &	0.206 &	0.069 \\
		$M_3$ & $\surd$ & $\surd$ &  &   &  0.063 &	0.640 &	0.084 &	0.199 &	0.210 &	0.077 \\
		$M_4$ & $\surd$ & $\surd$ & $\surd$ &  &   0.068 &	0.653 &	0.092 &	0.202 &	0.217 &	0.079 \\
		$M_5$ & $\surd$ & $\surd$ & $\surd$ & $\surd$  & \textbf{0.081} & \textbf{0.665} & \textbf{0.106}  & \textbf{0.209} & \textbf{0.228} & \textbf{0.083}  \\
% 		\hline
% 		\hline
\bottomrule
	\end{tabular}
}
\caption{Analysis of ablation study on IMDB-B and AnchorQuery-25K datasets. IGC, CGC, AAG and NPL correspond to Instance-level Graph Contrast, Cluster-level Graph Contrast, Adaptive Affinity Graph and Neighbor-aware Pseudo Label, respectively.\label{tab:ab}}
\end{table}

\smallskip
\noindent\textbf{Effect of Adaptive Affinity Graph.}
The difference between $M_4$ and $M_3$ lies in whether to use the adaptive affinity graph or not. It can be observed that $M_4$ achieves better results than $M_3$ on both datasets, which proves that extra neighbor information provided by the affinity graph is beneficial to learning better cluster assignments. Moreover, compact neighbors can better increase the intra-cluster information, and better serve the clustering process.

\smallskip
\noindent\textbf{Effect of Neighbor-aware Pseudo Label.}
Comparing the results of $M_4$ and $M_5$, we can draw the conclusion that the neighbor-aware pseudo-labeling mechanism is essential and can further improve the clustering accuracy by $1.0\%$ on both datasets, which demonstrates that effective supervision signals provided by pseudo-labels via cluster assignments can in turn optimize the representation learning.

\begin{figure}[t]\small
\centering
\subfigure{

\centering
\includegraphics[width=0.22\textwidth]{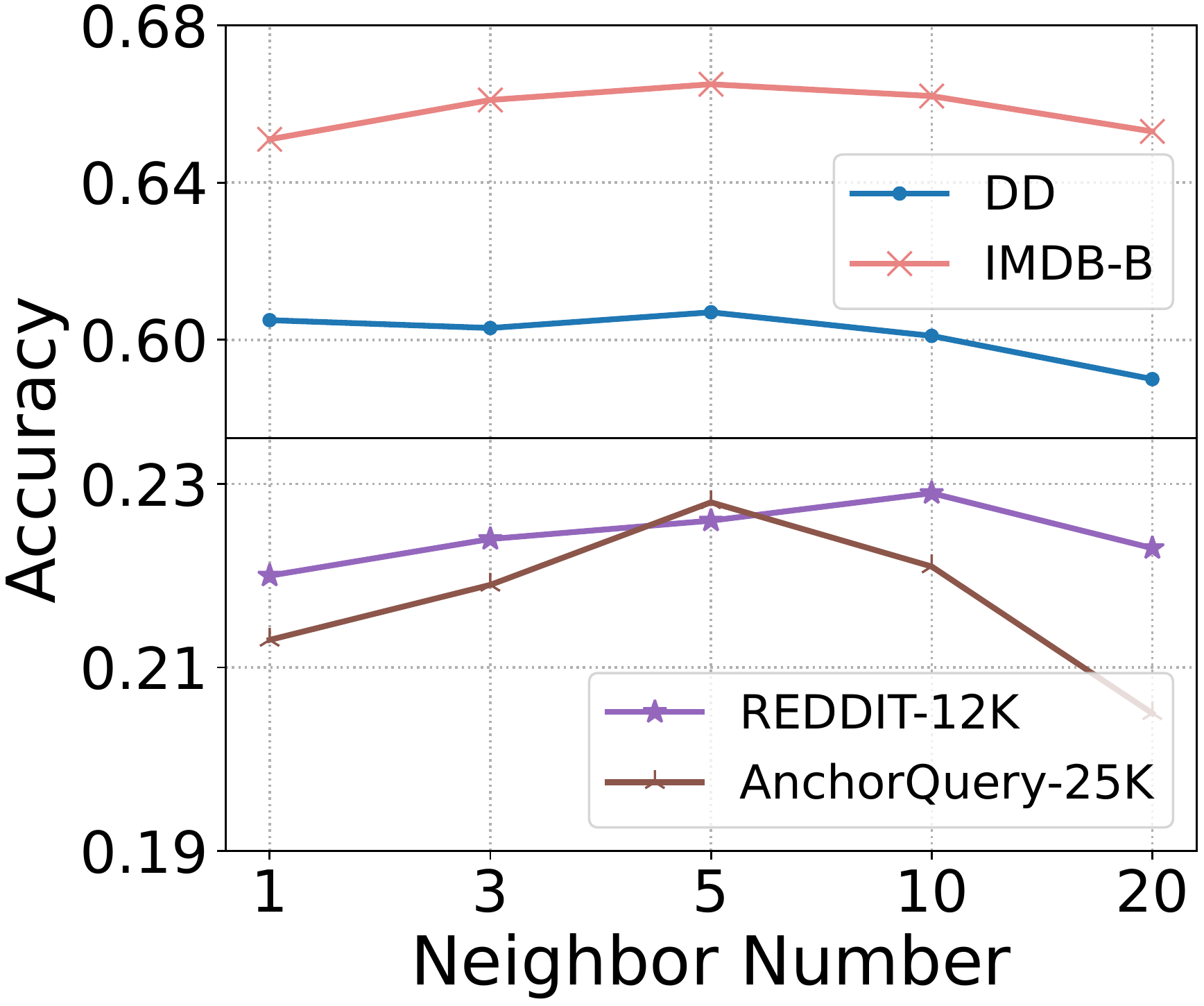}
}
\subfigure{

\centering

\includegraphics[width=0.22\textwidth]{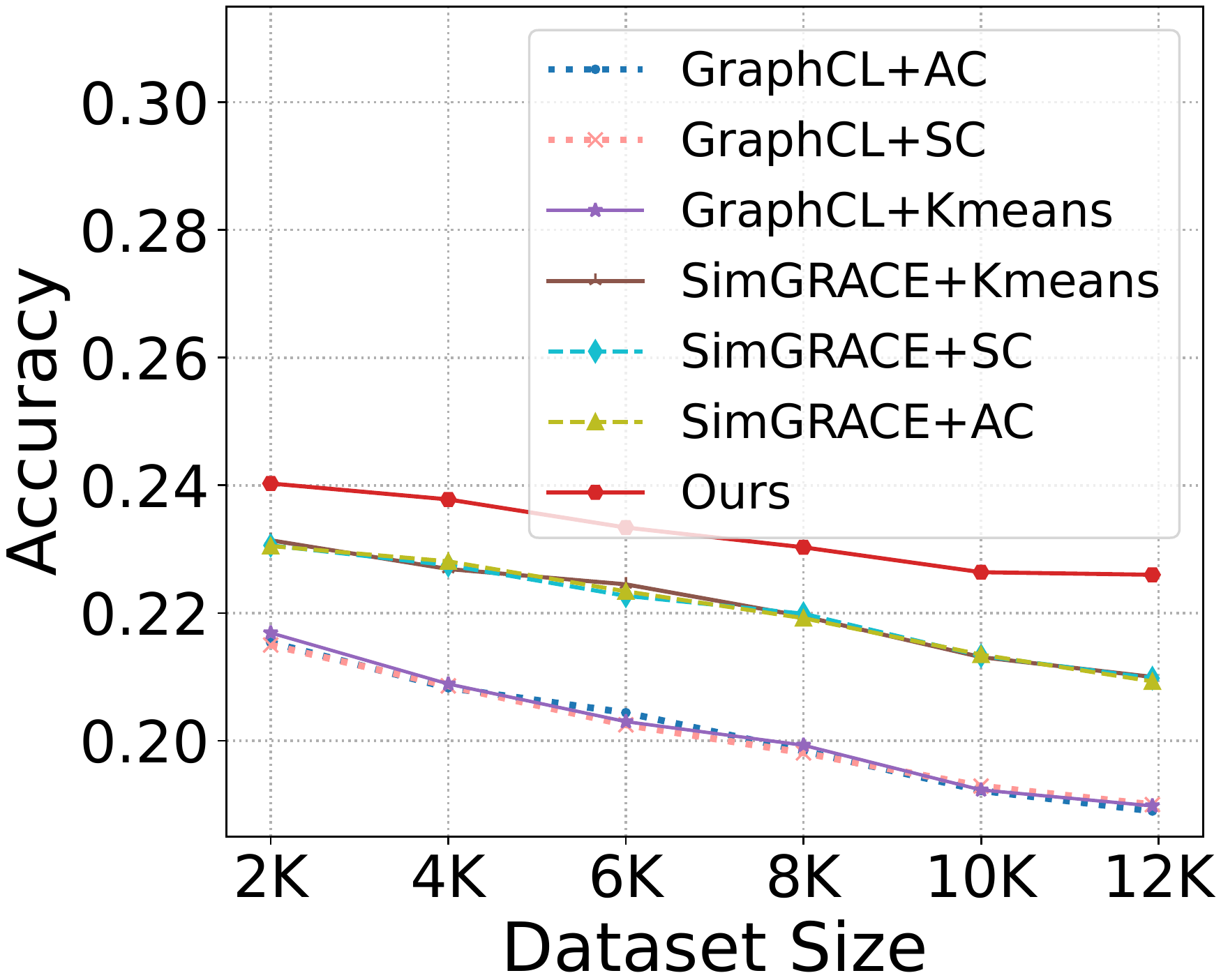}
}
\caption{Performance \textit{w.r.t.} neighbor number on four datasets and dataset size on REDDIT-12K.}
\label{fig:sensitivity}
\end{figure}

\subsection{Sensitivity Analysis}

\smallskip
\noindent\textbf{Analysis of Neighbor Number.} 
We first analyze the effect of the neighbor number $k$ of the affinity graph on four datasets. As shown in Figure \ref{fig:sensitivity} (Left), the performance improves gradually when the number of neighbors increases from $1$ to $5$. We deem that this improvement is mainly from richer intra-cluster information provided by more neighbors, which increases intra-cluster similarity while decreasing the false-negative samples in contrastive learning. Nevertheless, the too large neighbor number could hurt the performance owing to a decrease in the probability of two samples belonging to the same category as their distance increases, which brings the noise to contrastive learning.

\smallskip
\noindent\textbf{Effect of Cluster Sample Size.} 
Here we sample subsets of REDDIT-12K with various sizes. Specifically, we sample subsets with sizes from $2000$ to $10000$, while keeping the original sample proportion among different categories. We compare our model against six competitive baselines, which are combinations of methods based on graph representation learning $\{$GraphCL, SimGRACE$\}$ and cluster algorithms $\{$K-means, spectral clustering (SC) \cite{shi2000normalized}, agglomerative clustering (AC) \cite{gowda1978agglomerative}$\}$. The results are shown in Figure \ref{fig:sensitivity}  (Right). It can be observed that the performance of our \method{} drops more slowly than other methods with the growth of dataset size, 
indicating its robustness to various cluster sample sizes. Moreover, three cluster algorithms $\{$Kmeans, SC, AC$\}$ show similar trends based on the same graph representation.

\subsection{Analysis of Neighbor-aware Labeling Mechanism}

\begin{figure}[t]\small
\centering
\subfigure{

\centering
\includegraphics[width=0.22\textwidth]{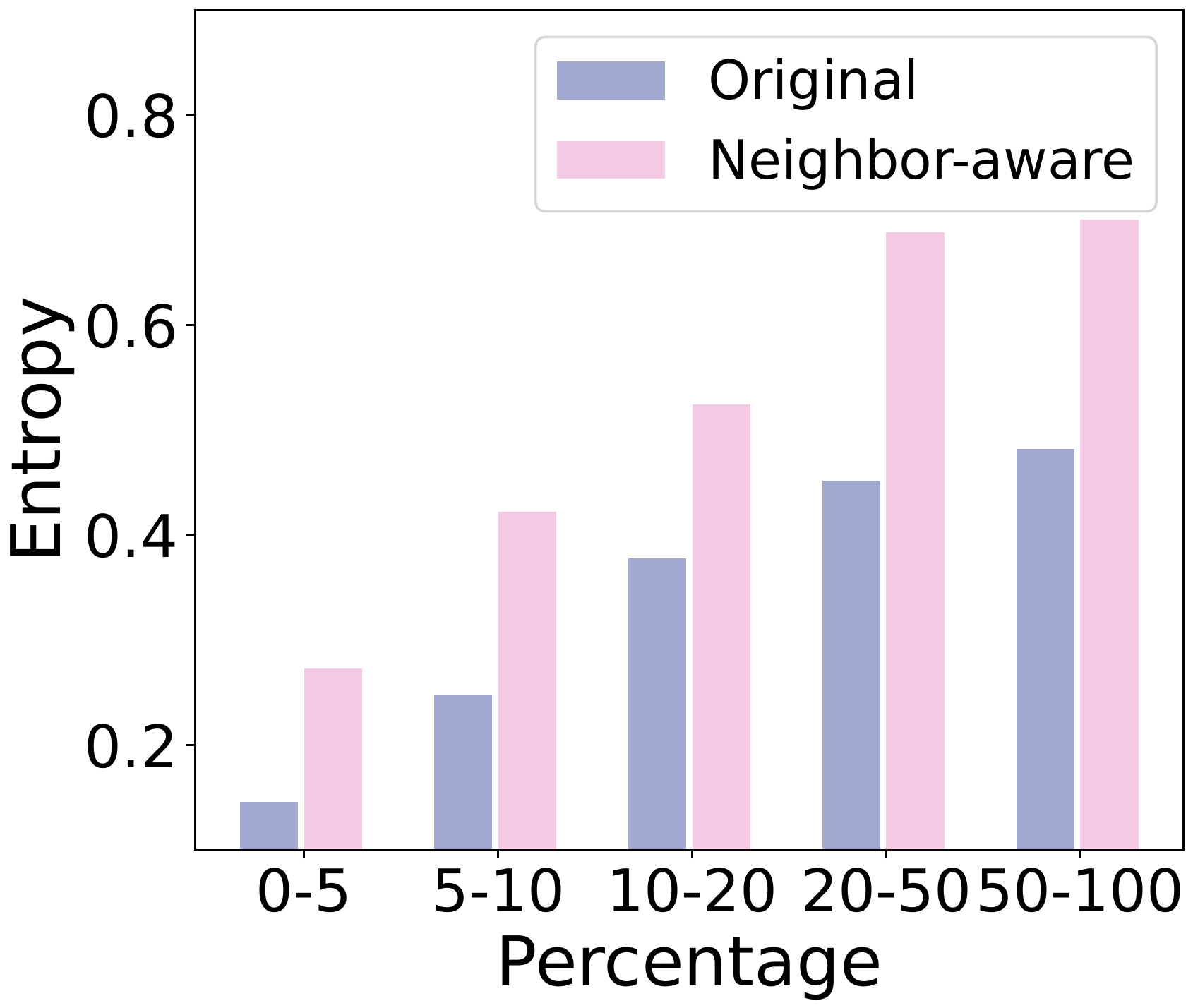}
}
\subfigure{

\centering
\includegraphics[width=0.22\textwidth]{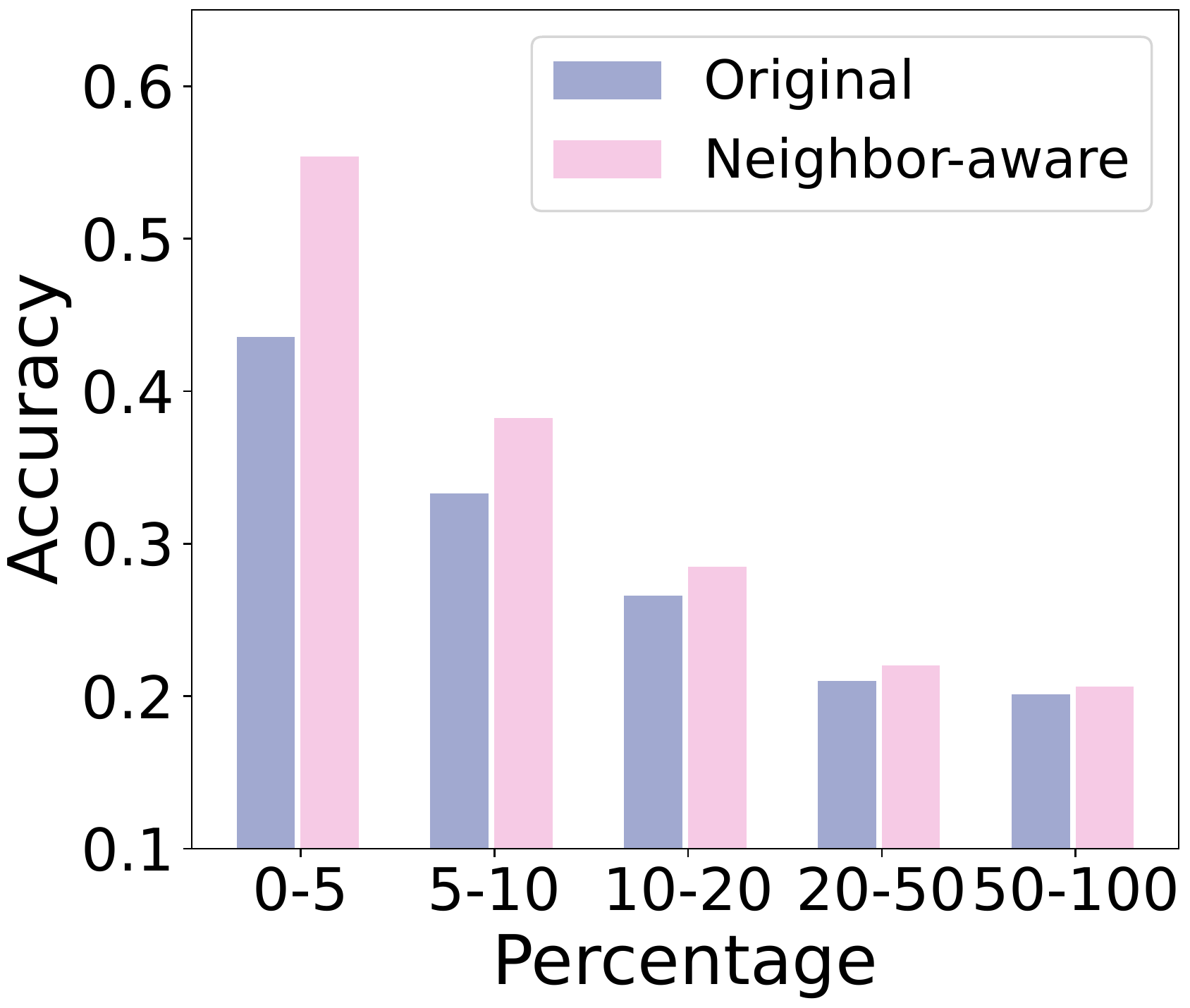}
}
\caption{Analysis of neighbor-aware labeling  mechanism on AnchorQuery-25K.}
\label{fig:neighbor_aware}
\end{figure} 

To look deep into the effect of neighbor information for pseudo-label selection, we report the distribution of entropy (left) and accuracy (right) of learned cluster assignments with respect to confidence degree in Figure \ref{fig:neighbor_aware}. Specifically, we consider the samples in five disjoint percentage intervals, i.e., $[0, 5)$, $[5, 10)$, $[10, 20)$, $[20, 50)$, $[50, 100)$, according to the confidence of original (neighbor-aware) assignments. $[0, 5)$ means top-$5\%$ confidence. Then we compute the averaged entropy and assignment accuracy of each group. As we can see, the averaged entropy of
original assignments is less than $0.5$, which means the original assignments produced by the well-trained model are close to one-hot vectors. However, their accuracy is not high, even though the model assigns a class with a probability close to one. Thus, we leverage the neighbor information from the affinity graph to relieve assignment bias. It shows that neighbor-aware assignments achieve higher accuracy compared with original assignments, at high confidence intervals, which proves the effectiveness of the neighbor-aware labeling mechanism.

\subsection{Discussion on Graph Augmentation}

\begin{table}[!t] 
\footnotesize
\centering
\tabcolsep=4.5pt
\resizebox{0.48\textwidth}{!}{
\centering
	\begin{tabular}{@{\extracolsep{-0.1cm}}cccccccccc@{}}
% 		\hline
% 		\hline
\toprule
		\multirow{2}{*}{Dataset}  & Aug  &
		ND & EP & SG & AM & \multicolumn{4}{c}{Random Selection} \\
		\cmidrule{3-6} \cmidrule{7-10}
		& Ratio & \multicolumn{4}{c}{0.1} & 0.05 &	0.1&	0.2&	0.4 \\
		 \midrule 
% 		\hline
% 		\hline
        DD & GraphCL  & 0.562&	0.563&	\underline{0.579}&	0.571&	0.562&	0.573&	0.571&	0.559   \\
		  & GLCC&	0.581&	0.567&	\underline{0.619}&	0.592&	0.611&	0.607&	0.602&	0.595  \\
		  \midrule
		  IMDB-B  &  GraphCL&	\underline{0.545} &	0.544&	\underline{0.545}&	0.542&	0.541&	0.545&	0.542&	0.538\\
		       & GLCC&	\underline{0.663}&	0.642&	0.595&	0.662&	0.648&	0.665&	0.646&	0.624\\
		       \midrule
		RDT-12K     & GraphCL&	0.183&	\underline{0.187}&	0.175&	0.179&	0.177&	0.181&	0.175&	0.166\\
		 &  GLCC&	0.220&	\underline{0.235}&	0.223&	0.232&	0.222&	0.226&	0.219&	0.198\\
		 \midrule
		 AQ-25K     & GraphCL&	0.187&	0.198&	\underline{0.203}&	0.189&	0.199&	0.201&	0.197&	0.161\\
		 &  GLCC&	0.193&	0.219&	\underline{0.234}&	0.203&	0.217&	0.228&	0.212&	0.168\\
		\bottomrule
% 		\hline
% 		\hline
	\end{tabular}
}
\caption{Analysis of Graph Augmentation. ND, EP, SG, and AM correspond to node dropping, edge perturbation, subgraph, and attribute masking, respectively. \label{tab:aug}}
\end{table}

Here we study the effect of various graph augmentation strategies and perturbation ratios on contrastive graph-level clustering. We compare our model \method{} with GraphCL~\cite{you2020graph} on four datasets. According to the results shown in Table \ref{tab:aug}, the following observations can be derived: (i) Graph augmentation is critical to our framework. For various datasets, different augmentation strategies exhibit pivotal effects, meanwhile the random selection of four strategies achieves stable performance 
across various datasets.
(ii) Without collaborative learning of representation learning and clustering, GraphCL cannot obtain much performance gain from different augmentation strategies. Moreover, we find that more advanced augmentation methods, like RGCL~\cite{li2022let} based on invariant rationale learning, fail to perform well in graph-level clustering, from Table \ref{tab:result}.
(iii) The appropriate perturbation ratio can achieve optimal cluster performance. These observations provide the enlightenment that further exploration of graph augmentation strategies, based on our joint framework, may further enhance graph-level clustering.

\section{Conclusion}
\label{sec::conclusion}

In this paper, we introduce a general framework \method{} for graph-level clustering given multiple graphs, which captures multi-granularity information to provide a global characterization of graph instances. We first construct an adaptive affinity graph to link semantically similar samples, while then introducing instance- and cluster-level contrastive learning based on the affinity graph. Moreover, we predict neighbor-aware pseudo-labels to optimize the representation learning process instead. Extensive experiments on a range of well-known benchmark datasets prove the effectiveness of the \method{} for graph-level clustering. 
% Our future works will further extend our framework to other domains such as protein clustering and genome analysis in bioinformatics.

\section*{Acknowledgments}
This paper is partially supported by grants from the National Key Research and Development Program of China with Grant No. 2018AAA0101902 and the National Natural Science Foundation of China (NSFC Grant Number 62276002).

% \balance
% Use \bibliography{yourbibfile} instead or the References section will not appear in your paper
\bibliography{aaai23}

% \section{Acknowledgments}

\end{document}